\documentclass{article}
\usepackage[utf8]{inputenc}
\usepackage{amsmath}
\usepackage{amssymb}
\usepackage[affil-it]{authblk}
\usepackage{graphicx}

\begin{document}

\title{A Deep Learning Approach to the Prediction of Drug Side--Effects on Molecular Graphs}
\author[2,3,*,c]{Pietro Bongini}
\author[1,*]{Elisa Messori}
\author[2]{Niccol{\`o} Pancino}
\author[1]{Monica Bianchini}
\date{}

\affil[1]{Department of Information Engineering and Mathematics, University of Siena, 53100, Siena, Italy}
\affil[2]{Department of Information Engineering, University of Florence, 50139, Florence, Italy}
\affil[3]{Department of Computer Science, University of Pisa, 56124, Pisa, Italy}

\affil[c]{Corresponding author: pietro.bongini@unifi.it}
\affil[*]{Contributed equally}

\maketitle

\begin{abstract}
Predicting drug side--effects before they occur is a key task in keeping the number of drug--related hospitalizations low and to improve drug discovery processes. Automatic predictors of side--effects generally are not able to process the structure of the drug, resulting in a loss of information. Graph neural networks have seen great success in recent years, thanks to their ability of exploiting the information conveyed by the graph structure and labels. These models have been used in a wide variety of biological applications, among which the prediction of drug side--effects on a large knowledge graph. Exploiting the molecular graph encoding the structure of the drug represents a novel approach, in which the problem is formulated as a multi--class multi--label graph--focused classification. We developed a methodology to carry out this task, using recurrent Graph Neural Networks, and building a dataset from freely accessible and well established data sources. The results show that our method has an improved classification capability, under many parameters and metrics, with respect to previously available predictors.
\end{abstract}

\section{Introduction}
\label{sec:introduction}
Drug Discovery is a fundamental but expensive process to make new pharmacological products available for healthcare \cite{ADR_drug_discovery}. Detecting and identifying Drug Side--Effects (DSEs) is mandatory to ensure that only safe drugs enter the market. DSEs have high costs for public healthcare \cite{ADR_costs}, and cause a high number of hospitalizations every year \cite{ADR_numbers}, a trend which is constantly increasing, also due to the increasing use of prescription drugs \cite{ADR_increase}. Predicting DSEs automatically in silico, before submitting drug candidates to clinical trials, would represent a fundamental improvement for drug discovery processes, cutting their costs in terms of time and money \cite{Predictor_Mizutani}.
\\
Automatic DSE predictors have traditionally relied on euclidean data representations \cite{Predictor_Mizutani} or drug similarity \cite{Predictor_Zhang}. In the last decade, we have seen an evolution towards Machine Learning (ML), with methods based on Random Forests \cite{Predictor_RF}, Support Vector Machines \cite{Predictor_Metabolic}, and Clustering \cite{Predictor_Drugclust}. Substantial improvements have been brought by the use of Deep Learning (DL) techniques which integrate heterogeneous data sources \cite{Predictor_Deepside}. Indeed, the number and variety of features to be used for prediction have steadily increased, as DSEs are complex biological phenomena involving many metabolic and genetic mechanisms \cite{ADR_Metabolomics}.
\\
Since their introduction, Graph Neural Networks (GNNs) \cite{GNN_model} have represented a very powerful model \cite{GNN_review_properties} for the prediction \cite{GNN_ppi} and generation \cite{GNN_mg2n2} of graph--structured data, with a wide variety of applications in the biological domain \cite{GNN_biognn}. Their capability of processing relational data directly in graph form, with little information loss and high flexibility \cite{GNN_capabilities}, allows GNNs to be successfully applied to an enormous variety of different tasks involving graph--structured data \cite{GNN_review_applications}. As a consequence, an ever increasing number of models have been developed to improve the field and to adapt the base theory \cite{GNN_theory} to the various scenarios, like Graph Convolution Networks (GCNs) \cite{GCN_standard}, spectral GCNs \cite{GCN_spectral_Bruna} \cite{GCN_spectral_Defferrard}, GraphSAGE \cite{GraphSAGE}, GraphNets \cite{GraphNets}, Message--Passing Neural Networks \cite{MPNN}, and Graph Attention neTworks (GAT) \cite{GAT}, just to name the most important. The whole GNN family has been classified into categories in order to better navigate through the configurations and to better study their properties from a mathematical point of view. This latter idea has lead to interesting ways of measuring their power in order to maximize the theoretical capabilities of future models, either using unfolding trees \cite{GNN_capabilities} or Weisfeiler--Lehman tests \cite{GNN_power}. 
\\
GNNs have been also applied to a related but very different task: polypharmacy effect prediction, in which the objective is to determine if two compounds are likely to trigger adverse reactions if taken together. In this framework, two main approaches exist: the first predicts the interactions as edges in a knowledge graph that conveys metabolomics and interactomics information \cite{Polypharmacy_Decagon}, while the second exploits a GAT--based graph co--attention mechanism on the two molecular graphs of each pair of compounds, training the co--attention mechanism to estimate the likelihood of a polypharmacy effect between the two drugs \cite{Polypharmacy_Deac}.
\\
In this paper, the DSE prediction is addressed as a multi--class multi--label classification problem. As the drug structure can be efficiently encoded by a molecular graph, we exploit GNNs to learn and automatically predict DSEs based on the drug structure only. This substantially differentiates the methodology from the only other GNN--based DSE predictor we are aware of: DruGNN, which predicted DSEs on a large knowledge graph integrating drug features, gene features, gene--gene interactions, drug--gene interactions and drug--drug similarities \cite{GNN_DruGNN}. While being capable of better integrating information from heterogeneous domains, thanks to the properties of Composite Graph Neural Networks (CGNNs), DruGNN is not capable of exploiting the full molecular information, because the drug structure is encoded as a fingerprint vector. Using SMILES, like other non--graph based methodologies do, also implies a loss of information. Molecular graphs instead retain the full amount of structural information that can be associated to each drug compound.
\\
The main contributions of this paper are as follows.
\begin{itemize}
\item	A novel dataset of molecular graphs is introduced: it can be used for the prediction of DSEs with any predictor model that accepts the drug structure in input; molecular graphs can also be enriched with relevant chemical features of the compound;
\item	GNN--MGSEP (Graph Neural Network --- Molecular Graph Side--Effect Predictor), a GNN--based model for the prediction of DSEs is introduced and validated on the dataset presented above; the problem is tackled as a graph--focused multi--class multi--label classification, where DSEs represent the class labels of the molecular graphs;
\item	Usability of GNN--MGSEP is discussed, and potential applications of the method in the real--world are introduced.
\end{itemize}
The rest of the paper is organized as follows: Section \ref{section:method} explains the methodology, describes the GNN model used for carrying out the predictions, and defines the experimental setup; Section \ref{section:results} presents and details the relevant experimental results for the validation of the method; Section \ref{section:discussion} discusses the relevance of the results, shows the prediction performance in comparison with other available methods, and describes the usability of the method; Section \ref{section:conclusions} draws conclusions on the work presented, discusses its impact and significance, and introduces possible future research.

\section{Materials and Methods}
\label{section:method}
\subsection{The Graph Neural Network model}
The Graph Neural Network (GNN) model is able to directly process a vast class of graphs by implementing a function $\tau(\mathbf{G},\,n) \in \mathbb{R}^m$ that maps each graph--node pair (in which the node belongs to the graph) into a $m$--dimensional Euclidean space. Therefore, the domain ordinarily considered is the set $\mathcal{D} = \mathcal{G}\times \mathcal{N} $ of graph--node pairs \cite{GNN_model}. GNNs can operate with most types of graphs --- acyclic, cyclic, directed, undirected ---, which allows to employ molecular graphs as input for such model without losing any topological information contained in chemical structure of the molecules. 
In our setting, we modelled a prediction problem as a supervised learning task: therefore, the learning set can be defined as in Eq. (\ref{eq:task}).
\begin{equation}
\label{eq:task}
   \mathcal{L}=\{(\mathbf{G}_i, n_{i,j},\mathbf{t}_{i,j}) \mid \mathbf{G}_i = (\mathbf{N}_i, \mathbf{E}_i) \in \mathcal{G};\, n_{i,j} \in \mathbf{N}_i;\, \mathbf{t}_{i, j} \in \mathbb{R}^m,\, 1\leq i \leq p, \, 1 \leq j \leq q_i\}
\end{equation}
where $n_{i, j} \in \mathbf{N}_i$ denotes the $j$--th node in the set $\mathbf{N}_i \in \mathcal{N}$ and $\mathbf{t}_{i, j}$ is the target array associated to $n_{i, j}$; $p$ is such that $p\leq |\mathcal{G}|$ and $q_i \leq |\mathbf{N}_i|$ holds for the number of supervised nodes $q_i$ in $\mathbf{G}_i$. However, it is evident that our application is \textit{graph--focused}, since the side--effects are relative to the whole molecule (there is no direct dependence on the individual nodes of the graph, which contribute equally to the output). \\
State functions and output functions are both computed via Multi--Layer Perceptrons (MLPs): for the state network, we employed the SeLU function with default parameters proposed in \cite{Activation_Selu} and the LeCun Normal initialization for the weights, while in the output network the sigmoid function and the Glorot normal initialization were used.\\
Our implementation is based on \textit{GNNKeras} \cite{GNN_keras}, a flexible tool which allows the construction of a large subclass of GNNs. The model, therefore, implements the state function as in Eq. (\ref{eq:state}), where the state of node $n$ at iteration $t$, $x_{n}^{t}$ is described in function of the states of the node itself and its neighbors at the previous iteration $t-1$, and the labels of $n$, its neighbors and the edges connecting them to $n$. The function $Ne(n)$ returns the set of all the neighbors of each node $n$, while the state updating function $f_w$ is implemented by the state MLP network:

\begin{equation}
\label{eq:state}
   x_{n}^{t} = f_w ( x_{n}^{t-1},\ l_{n},\ a \sum_{m \in Ne(n)} ( x_{m}^{t-1},\ l_{m},\ e_{m,n} ) )
\end{equation}

\noindent
Since the problem is graph--focused, the output function is described as in Eq. (\ref{eq:output}). The set $N_{out}$ contains all the output nodes, in our case all the nodes of the graph $N_{out} = N$. The output function $g_{w}$ is implemented by the output MLP network.

\begin{equation}
\label{eq:output}
   y_{G} = \frac{1}{|N_{out}|}\sum_{n \in N_{out}}g_w(x_{n}^{K},\ l_{n})
\end{equation}

\subsection{Data and pre--processing}
\begin{figure*}[!t]
\centering
\includegraphics{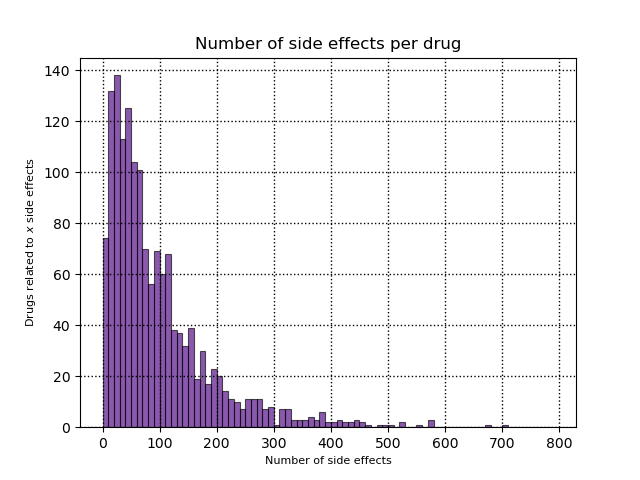}
\caption{The number of side--effects per drug follows a non uniform skewed distribution. Most drugs have few DSEs, while few drugs are associated to a large number of DSEs. This causes unbalancement in the class distributions which can lead to a bias in the model. \label{histogram}}
\end{figure*}

The only data required in this framework is the drug chemical structure and some known drug side--effect associations. To retrieve such data, we used the public database SIDER \cite{Dataset_Sider_1}, which contains data on 1430 drugs, 5880 Adverse Drug Reactions (ADRs), and 140,064 drug--ADR pairs \cite{Dataset_Sider_2}. We referred to the "stereo" version of the STITCH compound identifier \cite{Dataset_Stitch}, since it can be used as a key on the public database PubChem \cite{Dataset_Pubchem}.\\
Before retrieving a graph representation of drug chemical structures, the SIDER database needed to undergo a pre--processing procedure in order to filter out duplicates of drug--ADR pairs, which occur inevitably due to the presence of both lowest level terms (LLTs) and primary terms (PTs) for side--effects: we filtered out all associations referred to LLT side--effects, so that our dataset was made only of drug--ADR pairs in which side--effects are expressed via a PT. In such a way, we are not working with duplicate pairs and the learning procedure is not negatively affected by unnecessary data. For the same reason, we applied a constraint on the side--effect occurrences and we decided to filter out ADRs with less than 5 occurrences in the dataset. Through these procedures, we went from having almost 309,000 associations to about 159,000 associations, with a significant reduction of side--effect multiplicity (from the original 4,251 to 2,055 side--effects with a number of occurrences equal to or greater than 5). In a further phase, we obtained a dataset of 157,000 associations, following the removal of 33 compounds who lacked any intramolecular bond and, therefore, could not be converted into a molecular graph with our current methodology. Moreover, Experiment B1 (see Section \ref{section:results}) was performed on a version of the dataset with an additional filter, i.e. by removing all drugs that were associated to either less than 5 or more than 400 side--effects; the distribution of number of side--effects associated for each drug is reported in Figure \ref{histogram}.

\noindent
Since we noticed that chemical elements do not have a uniform distribution in our dataset, we developed a grouping system so that similar elements could be viewed as the same by the GNN model, based on their physico--chemical properties (see Table \ref{tab:grouping}). This is fundamental because some elements have very few occurrences, meaning that the network cannot possibly learn to manage them as a standalone node type. Moreover, filling the network with too many node types would make the learning problem more difficult.

\begin{table}[!t]
\caption{Grouping of the elements employed in this work.}
\label{tab:grouping}
\centering
\begin{tabular}{|c||c||c||c|}
\hline
\textbf{Element group}	& \textbf{Element}  &\textbf{Element group}  &\textbf{Element}\\
\hline
1 & C & 9 & Br  \\
2 & N & 10 & Na, K, Li \\
3 & O & 11 & Ca, Mg, Ba, Sr\\
4 & S, Se &  12 & Co, Tc, Mn, Fe\\
5 & F & 13 & Au, Ag, Pt, Zn  \\
6 & P & 14 & B, Ge, In, Tl\\
7 & Cl & 15 & La, Gd\\
8 & I & & \\
\hline
\end{tabular}
\end{table}

\subsection{Molecular graphs}
\label{subsec:molecular_graphs}
We represented molecule chemical structures by employing graphs, which allows to minimize the loss of information that would be unavoidable when compressing the molecular graph into other data structures. In order to retrieve the graph representation of a specific molecule, some intermediate steps were necessary: by using PubChemPy we retrieved the SMILES \cite{Smiles} string associated to the compound, which was then transformed into a RWMol (i.e. an editable molecule class defined in the RDKit \footnote{RDKit: Open--Source Cheminformatics Software, by Greg Landrum. URL: https://www.rdkit.org/} library). The RWMol was subsequently exploited to build a NetworkX graph of the molecular structure. Finally, this latter graph was converted into a GraphObject, a Python object defined specifically to be used as a structured graph representation for GNNKeras \cite{GNN_keras}.\\
The graph representation used in this work is made of three different components:
\begin{itemize}
    \item the node matrix, where rows represent the chemical element (or chemical element grouping) of the specific node. The following general rule applies to the node matrix:
\begin{equation*}
    n_{ij}=\begin{cases}
    1 & \mbox{if }\mbox{the $i$--th node belongs to the $j$--th element group} \\
    0 & \mbox{otherwise}
\end{cases}
\smallskip
\end{equation*}
\item the edge list, made of arrays of length 6 that collect initial and final node of the edge, along with a label indicating the chemical bond it represents. More specifically, an edge label is composed of:
\begin{equation*}
    \lbrace n_h,\, n_k,\, \underbrace{b_1,\, b_2,\, b_3,\, b_4}_{{\substack{\text{one--hot array} \\ \text{expressing the bond type}}}} \rbrace
\smallskip
\end{equation*}
\noindent
where $n_h$ and $n_k$ are respectively the initial and final nodes and the remaining entries encode the bond type (namely single, double, triple, aromatic);
\item the target vector, employed to carry out supervised learning; it consists in a binary vector of 2055 entries (i.e., one entry per side--effect) such that
\begin{equation*}
    t_i=\begin{cases}
    1 & \mbox{if the drug can cause the $i$--th side--effect} \\
    0 & \mbox{otherwise}
\end{cases}
\end{equation*}
\end{itemize}

\section{Experimental Results}
\label{section:results}
\subsection{Experimental setting}
Side--effect prediction, in this work, is modelled as a classification problem. Therefore, it requires an adequate loss function to be optimized during learning. The best fitting choice in this scenario is the binary cross--entropy, that handles each DSE independently from the others by taking into consideration the entries of the network's output one by one. The loss function was optimized via the Adam optimizer (from Adaptative moment estimation), which has proved to be highly efficient, requires little memory, and is appropriate for problems with noisy and/or sparse gradients \cite{Adam}. We employed the default hyperparameter values for the Adam optimizer.\\
The results presented in the following are obtained by applying 5--fold cross--validation, to obtain an unbiased evaluation of GNN--MGSEP, by alternately training it on four folds and testing on the other one. The following settings differ from each other on various parameters, such as number of epochs, batch size and stopping criteria. These differences are presented and compared in Table \ref{tab:exp_setting}.

\begin{table}[!t]
\caption{Hyperparameters of the experiments discussed in this work. Exp. B and B1 were carried out using the same parameters but different datasets (see Section \ref{section:method}).}
\label{tab:exp_setting}
\centering
\begin{tabular}{|c||c||c||c||c|}
\hline
\textbf{Parameter} & \textbf{Exp. A}	& \textbf{Exp. B}	& \textbf{Exp. B1} & \textbf{Exp. C} \\
\hline
\textbf{Batch size} & 32 & 32 & 32 & 16\\
\textbf{Threshold loss}& 0.15 & 0.15 & 0.14	& 0.14\\
\textbf{Epochs} & 8000 & 10000 & 10000 & 7500 \\
\textbf{Patience}& 2000 & 2000 & 2000 & 1000 \\
\hline
\end{tabular}
\end{table}

Various metrics were employed in order to evaluate different aspects of the model's performance. One metric we considered is \textit{binary accuracy}, which is a simple computation of how often predictions match binary labels, expressed through a percentage. When using such metric, entries of target arrays are considered independent from each other, by performing an element--wise comparison between the predicted array and the desired output. Despite its relevance, binary accuracy does not provide enough information about the performance of the network, due to the nature of the target distribution. In fact, each target array consists of 2055 entries and each chemical compound in the filtered dataset causes $\approx97$ side--effects on average~\footnote{More precisely, the computation of the mean results in $96.744$ side--effects per compound.}, which accounts for only $4.71\%$ of the total number of target array entries: as a consequence, a high level of binary accuracy is not necessarily an indication of good network performance, since it could be reached even in the case in which the network's predictions consisted of vectors full of zeros. For this reason, we employed also the Area Under ROC Curve (AUC) and the Area Under Precision Recall Curve (AUPR): they are obtained by plotting the true positive rate against the false positive rate and by plotting the positive predicted value against the true positive rate, respectively.\\

\subsection{Results}
Table \ref{tab:results} shows the results obtained in each experiment: Exp. B1, which was carried out after a further filtering of the data, provided the best performance in terms of binary accuracy and AUC, while Exp. A resulted in a better AUPR.

\begin{table*}[!t]
\caption{Results for each experiment.}
\label{tab:results}
\centering
\begin{tabular}{|c||c||c||c||c|}
\hline
\textbf{Metric} & \textbf{Exp. A}	& \textbf{Exp. B} & \textbf{Exp. B1} & \textbf{Exp. C}\\
\hline
Binary accuracy (\%) & 95.16 $\pm$ 0.42 &  95.13 $\pm$ 0.43 &  95.25 $\pm$ 0.57 & 94.94 $\pm$ 0.34 \\
AUC (\%) & 86.13 $\pm$ 0.46 & 86.11 $\pm$ 0.95& 86.73 $\pm$ 0.55 & 85.86 $\pm$ 0.33 \\
AUPR (\%) & 29.13 $\pm$ 2.22 & 28.85 $\pm$ 1.84 & 28.54 $\pm$ 3.54 & 26.82 $\pm$ 1.67  \\
\hline
\end{tabular}
\end{table*}

The best three experiments underwent a further analysis: Table \ref{tab:further_analysis} reports positive predicted value, negative predicted value, specificity and sensitivity of experiments A, B and B1. It is worth noting how different the positive predicted value and the specificity are, compared to the negative predicted value and the sensitivity: therefore, it is possible to hypothesize that currently the GNN model, in the framework of side--effect prediction, is better at detecting negative associations with respect to positive ones due to the unbalanced distribution of drug side--effect associations.

\begin{table*}[!t]
\caption{Further analysis for the best experimental settings.}
\label{tab:further_analysis}
\centering
\begin{tabular}{|c||c||c||c|}
\hline
\textbf{Metric} & \textbf{Exp. A}	& \textbf{Exp. B} & \textbf{Exp. B1} \\
\hline
Positive predicted value (\%) & 47.56 $\pm$ 25.84 \textsuperscript{3} &  45.41 $\pm$ 24.22 &  45.12 $\pm$ 24.01 \\
Negative predicted value (\%) & 95.89 $\pm$ 4.44 & 96.00 $\pm$ 4.33 & 96.14 $\pm$ 3.63 \\
Specificity (\%) & 21.32 $\pm$ 14.59 & 23.82 $\pm$ 15.32 & 20.68 $\pm$ 14.02 \\
Sensitivity (\%) & 99.05 $\pm$ 1.36 & 98.91 $\pm$ 1.45 & 98.86 $\pm$ 2.36 \\
\hline
\end{tabular}
\\
\noindent{\footnotesize{\textsuperscript{3}}This was calculated removing predictions that yielded no association between 19 molecules and the side--effects.}
\end{table*}

An analysis focused on the side--effects revealed that the relative frequency of each adverse reaction highly influences the ability of the model in detecting cases of positive associations regarding such side--effects. Table \ref{tab:sideff_analysis} shows such results, by considering the 10 most frequent and less frequent side--effects (DSEs) in our datasets and reporting the ratio between true positive predictions and the number of occurrences of such adverse reactions. The difference in the "detectability" of side--effects based on their number of occurrences in the dataset is clearly shown.

\begin{table*}[!t]
\caption{Influence of the relative frequency of each adverse reaction on the model ability in detecting positive associations.}
\label{tab:sideff_analysis}
\centering
\begin{tabular}{|c||c||c||c|}
\hline
\textbf{Metric} & \textbf{Exp. A}	& \textbf{Exp. B} & \textbf{Exp. B1} \\
\hline
Most frequent DSEs  (\%) & 69.31 $\pm$ 11.28 &  85.59 $\pm$ 15.11 &  83.81 $\pm$ 15.61 \\
Less frequent DSEs (\%) & 1.50 $\pm$ 3.22 & 2.97 $\pm$ 4.52 & 5.68 $\pm$ 7.29 \\
Overall average (\%) & 11.18 $\pm$ 19.19 & 12.06 $\pm$ 21.71 & 10.90 $\pm$ 19.33 \\
\hline
\end{tabular}
\end{table*}

\section{Discussion}
\label{section:discussion}

\subsection{Relevance of the Results}
\label{subsec:relevance}
The experimental results described in Section \ref{section:results} demonstrate that the DSE prediction task can be effectively carried out exploiting just the drug structure (the molecular graph) as it is done by GNN--MGSEP. A similar task had been carried out with GNN (DruGNN) by exploiting a large knowledge graph containing as much as seven main information resources: drug structural fingerprints, drug chemical properties, gene molecular function ontology, genomic information, gene--gene interactions, drug--drug similarity and drug--gene interactions \cite{GNN_DruGNN}. The setup presented in this work is much simpler, with only drug structures needed for the prediction, yet the molecular graphs convey structural information which is very important to determine the drug functionality. This results in a simpler yet very efficient prediction framework as highlighted by the metrics.\\
For performance comparison, we will also take into account other predictors not based on GNNs, such as:
\begin{itemize}
    \item Pauwels \cite{Predictor_Pauwels}, which is a good structure--based baseline as it predicts DSEs based on the Sparse Canonical Correlation Analysis (SCCA) of structural fingerprints only;
    \item DrugClust \cite{Predictor_Drugclust}, which is a good predictor based on clustering and Gene EXpression (GEX) data;
    \item DeepSide \cite{Predictor_Deepside} that represents a more complex predictor based on deep learning and integrates heterogeneous data from different sources.
\end{itemize}
These methods together with DruGNN constitute a very good set of models allowing to evaluate the capabilities of GNN--MGSEP in comparison to what has been achieved so far. Of course, a direct comparison is not possible as all of these methods use different data types and have been therefore trained and tested on datasets of different nature. Yet, once assessed the differences on the types of data used, the size of the datasets, and the sets of DSEs which are predicted in each case, they allow to evaluate the placement of our method with respect to other predictors.\\
The number of drugs taken into account, the number of predicted side--effects, as well as the information used and the method behind each predictor are described in Table \ref{table:predictors}.

\begin{table*}[!t]
\caption{Comparison of data and methodology for each predictor. SF stands for Structural Fingerprints, MG for Molecular Graphs, GO for Gene Ontlogy data, CF for Chemical Features, DPI for Drug--Protein Interactions, DDS for Drug--Drug Similarity, PPI for Protein--Protein Interactions.}
\label{table:predictors}
\centering
\begin{tabular}{|c||c||c||c||c|}
\hline
\textbf{Predictor}	& \textbf{Num. Drugs}	& \textbf{Num. DSEs} & \textbf{Method} & \textbf{Data Types}\\
\hline
Pauwels\cite{Predictor_Pauwels}	& 888 & 1,385 & SCCA & SF\\
DrugClust\cite{Predictor_Drugclust}	& 1,080 & 2,260 & Clustering & CF+DPI+GEX\\
DeepSide\cite{Predictor_Deepside} & 791 & 1042 & MLP & GEX+GO+SF\\
DruGNN\cite{GNN_DruGNN} & 1,341 & 360 & CGNN & CF+SF+GO+
PPI+DPI+DDS\\
\textbf{GNN--MGSEP} & 1,397 & 2,055 & GNN & MG \\
\hline
\end{tabular}
\end{table*}

As demonstrated by DeepSide and DruGNN, integrating an increasing amount of heterogeneous information has been the key for improving DSE predictors so far. In particular, the former was one of the first DL approaches to the problem, while the latter introduced the use of GNNs for analyzing the knowledge graph of DSEs. GNNs though can also be exploited to analyze the structure of each molecule, since molecules are naturally represented by graphs. Moreover, molecular graphs convey the full structural information of the molecule more efficiently with respect to the structural fingerprints, which are widespread in this field. This results in comparable performance between DeepSide, DruGNN and our GNN--MGSEP, which uses a much simpler load of information consisting only of the molecular graph of each compound, therefore representing a very easy to use predictor with respect to the other two. The performance of each model is reported in Table \ref{table:comparison}.

\begin{table}[!t]
\caption{Comparison of prediction performance with respect to the other described methodologies. Please notice that each predictor was trained and tested on its own dataset, with different data types, number of drugs, and number of DSEs compared to the others. These are reported in Table \ref{table:predictors} and should be taken into account when comparing the performance of the predictors. Three metrics are considered for the evaluation: Binary Accuracy, AUC (Area Under ROC Curve), and AUPR (Area Under Precision--Recall curve). Only the metrics proposed by the respective authors are reported for each predictor, as different problem formulations do not always allow to use the same metrics.}
\label{table:comparison}
\centering
\begin{tabular}{|c||c||c||c|}
\hline
\textbf{Predictor}	& \textbf{Binary Accuracy}	& \textbf{AUC} & \textbf{AUPR}\\
\hline
Pauwels\cite{Predictor_Pauwels}	& - & $89.32\%$ & - \\
DrugClust\cite{Predictor_Drugclust}	& - & $91.38\%$ & $33.36\%$\\
DeepSide\cite{Predictor_Deepside} & - & $87.70\%$ & - \\
DruGNN\cite{GNN_DruGNN} & $86.30\%$ & - & - \\
\textbf{GNN--MGSEP} & $95.25\%$ & $86.73\%$ & $29.13\%$ \\
\hline
\end{tabular}
\end{table}

\subsection{Usability}
\label{subsec:usability}
The usability of GNN--MGSEP is very easy thanks to the low amount of information it needs to accurately predict the occurrence of side--effects: for estimating the probable DSEs of a new drug, it will be sufficient to submit its molecular graph to the model. There is no need to retrain the model every time that a new drug is introduced in the dataset. Yet, when a significant amount of new drugs becomes available together with their labels (known occurrence of side--effects) a retraining will improve the model performance for future predictions. Our model is lightweight and training it does not require heavy amounts of resources: in our experiments we just used a commercial laptop even without a GPU. Once the model is trained, obtaining a prediction with GNN--MGSEP is even more lightweight as there is no need to load the whole dataset of molecular graphs. Our model can therefore be used as a simple screening service to predict the occurrence of side--effects on massive amounts of molecular graphs, in the very early stages of a drug discovery pipeline.\\

\subsection{Future Developments}
\label{subsec:future}
In the future, the model can be further developed in multiple directions. On the one hand, introducing a larger amount of drug examples could improve performance of GNN--MGSEP while retaining the same simplicity and lightweight style. On the other hand, the model can be refined by integrating heterogeneous data as it is the case for DeepSide and DruGNN. The more straightforward addition that could be made is constituted by the chemical features of drugs, that can be retrieved from PubChem and integrated inside the molecular graph. Other data, describing drug--protein interactions, metabolomics, gene expression, and ontologies, could be integrated as well, though this would imply a rethinking of the model to attach these pieces of information to a molecular graph.\\
An integration with other DL tools thought for drug discovery is also possible. As GNN--MGSEP is ideal for screening huge amounts of molecular graphs, it represents a very good model for processing the output of molecular graph generators based on DL, such as ChemVAE \cite{Generation_ChemVAE}, JTVAE \cite{Generation_JTVAE}, CCGVAE \cite{Generation_CCGVAE}, GraphVAE \cite{Generation_GraphVAE}, MolGAN \cite{Generation_MolGAN}, or the GNN--based MG\textsuperscript{2}N\textsuperscript{2} \cite{GNN_mg2n2}. These methods can in fact produce massive amounts of possible drug candidates, but often lack the ability of evaluating the possible DSEs of the generated compounds. A three--step chain can also be devised, in which the graph generator constitutes the first step, aimed at producing a large pool of possible drug candidates. The drug candidates could then be screened for their drug--likeness, retaining only compounds with a high QED score \cite{QED} or druggability score, which can be estimated with various methods, including deep learning predictors \cite{DL_pocket_scan, PockDrug, GlyPipe}. GNN-MGSEP could then screen the selected compounds, filtering out those with too many or too dangerous side--effects.

\section{Conclusion}
\label{section:conclusions}
This paper presented GNN--MGSEP, a new model for the prediction of drug side--effects based on the molecular graph that describes the drug structure. A dataset of molecular graphs and associated side--effects was built in order to train and test the model. The experimental results show that the model is capable of very good performance on the task of drug side--effect prediction. Exploiting only the molecular graph, it is able to obtain comparable performance, and in some cases even better performance, with respect to the state of the art methods in such task, which need large loads of information from heterogeneous sources to formulate their predictions --- even though a direct comparison is not possible due to the different nature of the data used by each predictor. The usability of GNN--MGSEP was discussed, highlighting the ease of use and lightweight training procedure of the model. Future directions of research are very promising, including the possibility of refining the predictions by integrating more data, and using GNN--MGSEP in a pipeline fully based on deep learning. In this latter framework, a graph generator outputs huge amounts of molecular graphs of possible drug candidates, which are subsequently screened for their drug--likeness using a dedicated model, and then for their side--effects using GNN--MGSEP.

\section*{Acknowledgments}
The authors would really like to thank Professor Franco Scarselli for the very wise insights on Graph Neural Networks he gave to them when talking about this project.

\bibliographystyle{ieeetr}

\end{document}